# Performance Assessment of different Machine Learning Algorithm for Life-Time Prediction of Solder Joints based on Synthetic Data

S. Muench[1], D. Bhat[1], L. Heindel[2], P. Hantschke[2,3], M. Roellig[1,3,*], M. Kaestner[2,3]
[1]Fraunhofer Institute for Ceramic Technologies and Systems IKTS, 01109 Dresden, Germany
[2]Technische Universität Dresden, Institute of Solid Mechanics, 01062 Dresden, Germany
[3]Dresden Center for Fatigue and Reliability (DCFR), 01062 Dresden, Germany
*Corresponding author. E-mail: Mike.Roellig@ikts.fraunhofer.de

**Abstract**

This paper proposes a computationally efficient methodology to predict the damage progression in solder contacts of electronic components using temperature-time curves. For this purpose, two machine learning algorithms, a Multilayer Perceptron and a Long Short-Term Memory network, are trained and compared with respect to their prediction accuracy and the required amount of training data. The training is performed using synthetic, normally distributed data that is realistic for automotive applications. A finite element model of a simple bipolar chip resistor in surface mount technology configuration is used to numerically compute the synthetic data. As a result, both machine learning algorithms show a relevant accuracy for the prediction of accumulated creep strains. With a training data length of 350 hours (12.5 % of the available training data), both models show a constantly good fitting performance of $R^2$ of 0.72 for the Multilayer Perceptron and $R^2$ of 0.87 for the Long Short-Term Memory network. The prediction errors of the accumulated creep strains are less than 10 % with an amount of 350 hours training data and decreases to less than 5 % when using further data. Therefore, both approaches are promising for the lifetime prediction directly on the electronic device.

## 1. Introduction

The increase of autonomously acting machines, as well as autonomous driving vehicles, raises the need for electronics which are reliable, trustworthy and functional under all circumstances. New approaches to monitor the condition of electronics can help to detect successive degradation and to evaluate fatigue at an early stage. Accordingly, prediction over a defined period of time is also possible, maintenance can be planned and components can be replaced before they fail. System failures caused by a lack of electronic functionality can thus be avoided. Maintenance strategies can be divided into three categories: proactive maintenance, in which components are replaced at regular intervals. This is usually performed in safety-relevant areas such as aviation technology. The second category is reactive maintenance, in which components are replaced only after they have failed. This is used, for example, for non-safety-relevant components in the automotive sector. The optimum between the two strategies in terms of acceptable maintenance costs and minimum failure costs is the condition-based maintenance. The latter requires continuous data acquisition and evaluation, ideally performed by the electronic unit to be maintained itself. For this purpose, data-driven models are of particular importance due to their ability to model highly complex relationships and provide computationally efficient predictions

It is well known that a large proportion of electronic component failures are related to temperature-induced damage to solder contacts. [1] But there are only a few publications on the reliability of solder contacts using machine learning (ML) techniques. Law and Azid [2] used an artificial neural network to correlate design parameters with the fatigue life. Lee et al. [3] predicted the residual ultimate lifetime (RUL) of solder under mechanical fatigue loads. However, effect of temperature on solder joint reliability is not completely captured by any approaches. Most importantly, there is a necessity to address the data requirement, complexity and limitations with data-driven methods in this domain. Therefore, in the following we would like to propose ML approaches that can be implemented directly in the electronic device, because they are computationally efficient and almost real-time capable.

The approaches will be investigated with respect to their suitability for the prediction of the damage progress in solder contacts and to the amount of necessary training data. The solder contacts of a simple bipolar component are used as the object of investigation, because their damage behavior is quite well understood and, on the other hand, there is good transferability to higher-order quad flat no leads Packages (QFNs) in terms of solder contact geometry and damage behavior.

## 2. Materials and methods

Since temperature sensors are already installed in many electronic applications, a temperature-time profile can be chosen as input variable for the data-driven approach. In the future, it will also be possible to choose substitute variables that correlate with temperature, such as current, voltage, power or energy consumption. The typical indicators for quantifying this thermal-mechanical fatigue are the accumulated creep strain or the dissipated creep energy density during a temperature cycle [4]. Accordingly, either the strain approach or the energy approach can be used for a lifetime prediction [5]. In this contribution, the accumulated creep strain is used, which can then be utilized with a lifetime law corresponding to the specific solder alloy and the encountered failure mode to calculate the RUL. Therefore, the accumulated creep strain $\varepsilon_{\text{creep}}^{\text{acc}}$ becomes the target value of the approach. The procedure is shown in Figure 1.

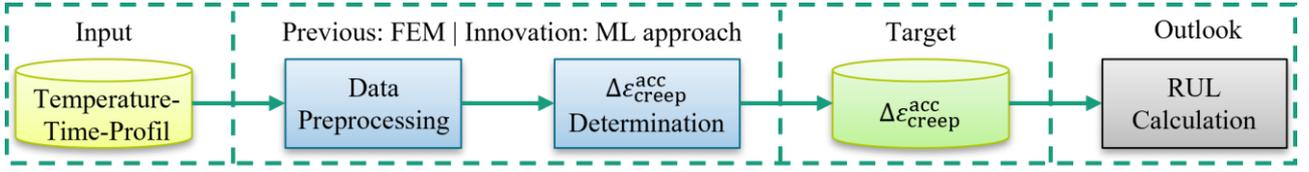

**Figure 1:** Procedure for determining creep strain increments from temperature profile data

For the investigation of the ML approaches, synthetic data were used. A realistic temperature profile for automotive applications serves as input. The resulting creep strain increments to be trained and predicted are calculated using a finite element analysis (FEA). Based on this data the ML approaches, namely a Multilayer Perceptron (MLP) and a Long Short-Term Memory (LSTM) network will be trained and validated.

**2.1. Synthetic data creation**

**2.1.1. Temperature-time-profile generation**

In order to examine the amount of necessary training data, a temperature profile of approximately 2780 h total duration is created which is later segmented into parts of different length. The total profile consists of 9910 consecutive half cycles, which themselves follow a typical exponential temperature curve. Equations (1) and (2) shows that the half cycle is completely defined by the start and target temperature and the exponent $a$.

$$T(t) = T_{\text{Target}} - \Delta T \exp(a\,t) \quad (1)$$

$$\text{with} \quad \Delta T = T_{\text{Target}} - T_{\text{Start}} \quad (2)$$

The largest temperature gradient $\dot{T}_{\max}$ in the half cycle corresponds to the slope of the function at the time $t = 0$. This condition can be used to obtain the definition of the exponent $a$ via the temperature gradient $\dot{T}$ from the derivative of the Equation (1):

$$\dot{T}_{\max} \stackrel{\text{def}}{=} \dot{T}(t = 0) = -a\,\Delta T \exp(0)\,. \quad (3)$$

The exponent $a$ is calculated by:

$$a = \frac{\dot{T}_{\max}}{\Delta T}. \quad (4)$$

Since the end temperature of a half cycle is also the start temperature of the following cycle, the parameters to be specified are reduced to the target temperature $T_{\text{Target}}$ and the maximal temperature gradient $\dot{T}_{\max}$. While the temperature step as well as the dwell time $d$, calculable via Equation (5) from the exponent $a$ assuming a residual deviation from the target temperature of 1 %, represent derived quantities.

**Table 1:** Range of the properties used to generate the synthetic temperature profile

| parameter | range | center | state |
|---|---|---|---|
| $T_{\text{Target}}$ [°C] | $-40\dots150$ | 25 | predefined |
| $\lvert\dot{T}_{max}\rvert$ [K/min] | $0.5\dots20$ | 7.5 | predefined |
| $\Delta T$ [K] | $-130\dots130$ | 0 | resultant |
| $d$ [min] | $1\dots165$ | -- | resultant |

$$d = \frac{1}{a} \ln\left(\left|\frac{0.01\,T_{\text{Target}}}{\Delta T}\right|\right). \quad (5)$$

The basic properties of the half cycle were predefined by 10000 normally distributed random numbers in the typical range for automotive applications, shown in Table 1.

**2.1.2. Finite element modelling**

The accumulated creep strains associated with the synthetic temperature profile were determined by FEA. The model of a bipolar chip resistor (CR) in surface mount technology (SMT) configuration used for this purpose is shown in Figure 2 and was already presented and verified in [6]. To give a short recapitulation we want to summarize the relevant main features here.

**Geometry:** Assuming an idealized state, the bipolar CR is simplified to a quarter model in order to reduce the required computing time. The specified geometric quantities correspond to an CR0805 and typical electronic manufacturing values for contact angle, total height, standoff and pad geometry. A printed circuit board (PCB) based on FR4-material was modeled as substrate.

**Material:** The thermo-mechanical fatigue of solder contacts is a time- and temperature-dominated process, so the corresponding time- and temperature-dependent modeling of the material behavior is significant. This is especially relevant for the viscoplastic behavior of the solder material (SnAg3.5). Due to the high homologous temperature of the solder materials, creep already occurs in the operating temperature ranges of electronic components. In the model, this behavior is represented by a secondary creep law, the hyperbolic sine creep law (Generalized Garofalo) according to Equation (6).

$$\dot{\varepsilon}_{\text{Creep}} = c_1 \sinh(c_2\,\sigma)^{c_3} \exp\left(\frac{c_4}{T}\right) \quad (6)$$

Here, the change in the equivalent creep strain with respect to time $\dot{\varepsilon}_{\text{Creep}}$ is computed by the equivalent stress $\sigma$, the temperature $T$ and the material dependent model parameters $c_1$ to $c_4$. These have already been characterized by Metasch et al. [7] in the range between -40 °C and 150 °C. Furthermore, the highly temperature-dependent behavior of the matrix resin of the PCBs must be considered. The material was already measured by Schwerz et al. [6] using a dynamic mechanical analyzer (DMA) and the necessary parameters were extracted from the measurement curves. The thermomechanical properties of the other materials involved are summarized in Table 2 with associated references.

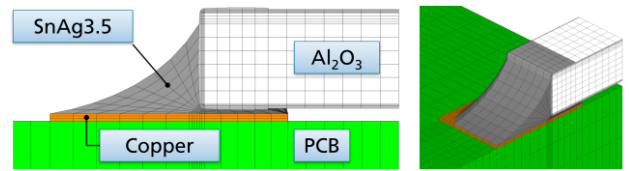

**Figure 2:** FEM model of the bipolar chip resistor used for computing the accumulated creep strains

**Table 2:** Overview of the material properties used for FEA based on [6]

| Material | CTE [ppm/K] | Elastic moduli [GPa] | Shear moduli [GPa] | Poisson Ratio [--] | Properties | Reference |
|---|---|---|---|---|---|---|
| Solder | 21.1 | 20.9 – 11.8 | - | 0.35 | elastic, viscoplastic | [7] |
| Ceramic | 7.0 | 330 | - | 0.30 | elastic | [8] |
| Nickel | 13.4 | 200 | - | 0.31 | elastic-plastic | [9] |
| Copper | 17.0 | 125 - 102 | - | 0.35 | elastic-plastic | [9] |
| FR4 | x,y 14.5 z 67.2 | x,y 14.3 - 3.8 z 5.7 - 1.5 | xy 8.7 - 5.9 xz, yz 3.8 - 2.6 | xy 0.11 xz, yz 0.44 | elastic, orthotropic | [10], DMA |

**Loading:** The previously generated realistic temperature profile is uniformly applied to all model nodes. Mechanical boundary conditions are only applied to avoid rigid body motion and to represent the symmetry-related model reductions. Boundary conditions in the context of a clamping of the PCB are not present.

**Evaluation:** Due to a well-developed meniscus with a relatively high solder volume content, the creep strains that occur are concentrated in the small standoff with a low solder volume. Accordingly, the first damage to SMT-mounted resistors also occurs in the standoff, as Schwerz et al. [6] has shown. Based on these results, the evaluation of the accumulated creep strains concentrates on the elements in the standoff region of the solder contact. After each calculated half cycle, the accumulated creep strain of these elements is extracted and calculated to a volume-averaged scalar quantity, see Equation (7):

$$\varepsilon_{\text{creep}}^{\text{acc}} = \sum_{i=1}^{n} \frac{\varepsilon_{\text{creep},i}^{\text{acc}}}{v_i}. \qquad (7)$$

For this, the element volumes $v_i$ of the Elements $i = 1 \rightarrow n$ are used in the reference configuration. The FE simulations were performed with the software ANSYS® Mechanical APDL, Release 18.2.

## 2.2. Data-driven approaches

For the application of data-driven approaches to the prediction of creep strain increments based on temperature profiles, two approaches are presented and compared in the following. Both approaches are implemented using the Python libraries Keras [11] and Tensorflow [12].

### 2.2.1. Multilayer Perceptron

A MLP is a subset of a deep feedforward artificial neural network. It is comprised of at least three layers, an input, a hidden, and an output layer. It has a fully connected structure, so every neuron is linked to all neurons in the subsequent layer. Because of the flexibility of adding hidden layers, the MLP is efficient in capturing complex non-linear relationships. It should be mentioned that an MLP determines an output value from a limited number of input values. The process could be called value-to-value [13, 14]. Figure 3 shows the MLP basic structure used in this work schematically.

In order to accurately predict output values from input data, the MLP network is parameterized using the back propagation algorithm, which is referred to as training the model. The target of this algorithm is to iteratively minimize a chosen loss function [15]. This is done by four main steps, which are: I) the weight initialization, II) the forward pass, III) the back propagation of the errors and IV) the weights update. For more information about how a MLP is trained, please see [13]. To quantify the errors in step II, a modified relative absolute error has been chosen as loss function with which importance is given to higher creep increments which have significant impact on predicted lifetime. The loss function is defined as

$$L_{\text{MLP}}(y, y^*) = \text{abs}\left(\frac{y - y^*}{y}\right), \qquad (8)$$

with a so-called true value $y$ and the predicted value $y^*$. For the special case considered in this contribution $y$ corresponds to creep strain increment computed by finite element solution for a given set of properties of a temperature half cycle and $y^*$ is the creep strain increment predicted by the databased approach. To initialize the weights at the first step, the He-Normal initialization is implemented for each layer [16].

### 2.2.2. Long Short-Term Memory network

The LSTM network [17, 18] is a recurrent neural network type that has been successfully used for a large variety of sequence modelling problems. It consists of LSTM blocks with a gated structure, which addresses the vanishing and exploding gradient problems of regular recurrent neural networks. In this contribution, the LSTM network was used to predict creep strain increments based on a temperature-time sequence. The process could be called sequence-to-value. For this specific application, different LSTM network architectures were tested and the best performing network structure is presented. It uses two LSTM blocks with a skip connection, depicted in Figure 4.

During model training, a squared error loss function

$$L_{\text{LSTM}}(y, y^*) = (y - y^*)^2 \qquad (9)$$

is deployed, where similar to Equation (8) $y$ corresponding to the true value and $y^*$ to the predicted value.

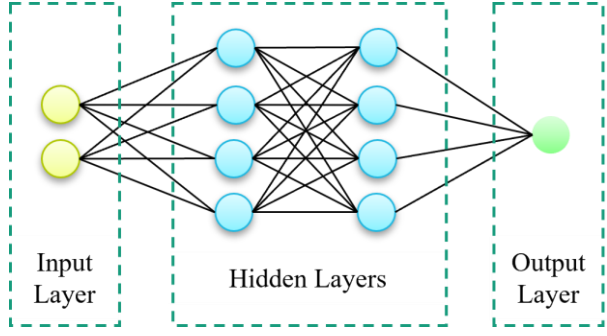

**Figure 3:** Multilayer Perceptron model

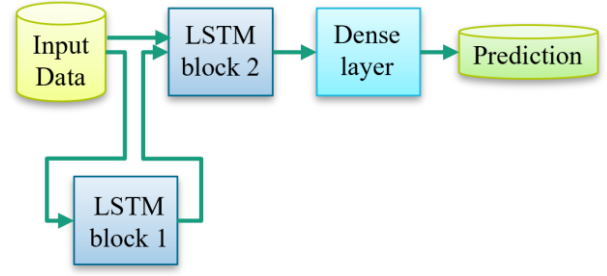

**Figure 4:** LSTM network architecture

### 2.2.3. Hyperparameters

All parameters that influence the training process, but are not part of the model itself, are called hyperparameters. It is quite understood that hyperparameters play a major role in the process of creating an accurate and robust ML model. It is also challenging to find the relevance of some hypermeters, because it is data dependent [19]. Here is a list of important hyperparameters that are considered in this article,

- Optimization algorithm: Algorithm to optimize the weights iteratively based on the calculated losses.
- Initial Learning rate: It is a measure which dictates the model update in response to error
- Dropout rate: A regularization parameter to reduce overfitting
- Batch size: Number of sub samples after which weights are updated in each epoch
- Epochs: Number of times training data is passed to the model
- Memory cells per LSTM block: A high number of memory cells leads to more complex LSTM models, which can achieve a higher prediction accuracy, but are also generally more likely to overfit and therefore require larger amounts of training data.
- sequence length: Determines the length of the sequence that is used as input data to the LSTM model.
- Sequence overlap: The overlap factor determines, by what fraction one sequence overlaps with its neighboring sequences.

## 3. Results
### 3.1. Data-understanding

Before correlating data, the first step is to understand the data. A special characteristic of the target variable, the increment of the accumulated creep strain, is that it covers a very large range of values from 8 E-14 to 6 E-3 for the given temperature profile. Since conventional loss functions consider the absolute error to judge and modify weights, this would mean that very large increments would have a high impact on the training, while small increments would be neglected. Therefore, to ensure that the particular approach chosen learns the given data uniformly, it is recommended to scale it by logarithm. For the MLP the natural logarithm according to Equation (10), and for the LSTM network the logarithm to the base 10 according to Equation (11) was used.

$$\Delta\varepsilon_{\text{creep}}^{\ln} = -\ln(\Delta\varepsilon_{\text{creep}}^{\text{acc}}) \qquad (10)$$

$$\Delta\varepsilon_{\text{creep}}^{\log} = -\log(\Delta\varepsilon_{\text{creep}}^{\text{acc}}) \qquad (11)$$

This reduces the value range from 11-decades to 1, see Figure 5. Please note that due to the negative sign in Equation (10), especially small creep strain increments now correspond to large values and vice versa. Once this transformation has been performed, the pair plot Figure A-1 is suitable for interpreting distributions and bivariate correlations.

The evaluation of the data-driven models should also be based on the necessary amount of training data. For this purpose, different sized portions between 3 % and 100 % of the synthetic training dataset are presented to the models. It is therefore important to check in advance whether the separated data sets have basically similar value ranges and distributions. As the boxplots Figure 6 a-e show, this condition is fulfilled. For all 5 diagrams, the mean values, median values and the area marked by the box between the 25th and 75th percentiles are almost congruent. There are some deviations in the whiskars, which indicate the minimum and maximum values of the underlying data. These extreme values are rare due to the normal distribution used to generate the data and therefore only occur when large numbers of training data are used.

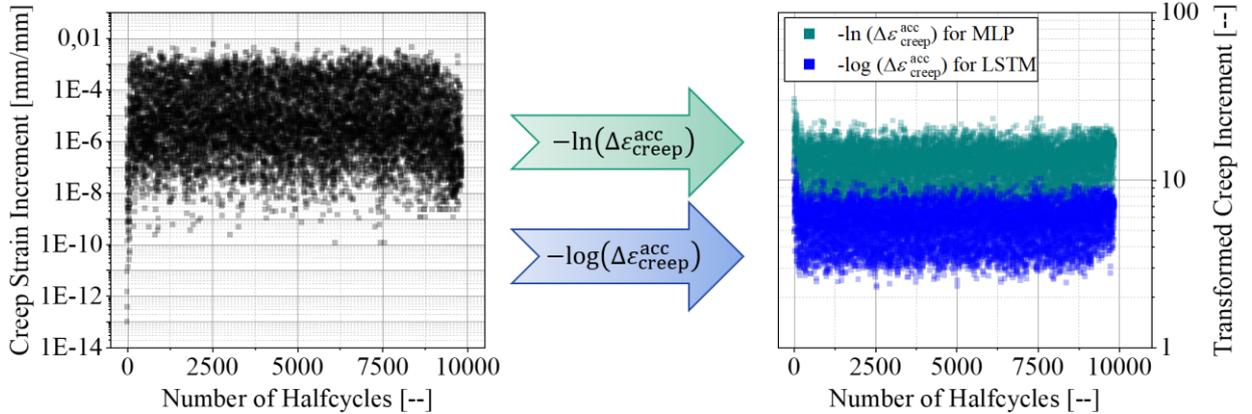

**Figure 5:** Reduction of the large value ranges through logarithmic scaling

### 3.2. Data preparation

For the usage in the data-driven approaches, the input data has been standardized (zero mean, standard deviation of 1) since input features are normally distributed [20]. No data scaling has been applied on output features of the MLP model, while the output data of the LSTM model has also been standardized. The data set is further split by 80-20 ratio to create a validation set, which is not used for training or updating the weights. It is purely to understand how well model is performing during the training.

While the MLP works with the parameters derived from the temperature profile per half cycle, the LSTM network is employed to predict the cumulated logarithmic creep strain increment caused by given temperature load sequences. These sequences of are extracted from the complete temperature load history of the finite element simulation using a specified sequence length and sequence overlap.

### 3.3. Network evaluation metrics

Before adapting the different hyperparameters of the approaches, evaluation metrics have to be defined. We use the coefficient of determination ($R^2$-score) between prediction and true values and a customized error function as shown in the Equation (12).

$$f_{\text{rel}}^{\text{ave}} = \frac{1}{n} \sum_{i=1}^{i=n} \left| \frac{\varepsilon_{\text{creep,pred},i}^{\text{acc}} - \varepsilon_{\text{creep,true},i}^{\text{acc}}}{\varepsilon_{\text{creep,true},i}^{\text{acc}}} \right| \quad (12)$$

This custom metric captures the average error between the predicted creep history and the FE solution, called the true solution, over the entire test dataset. To calculate, the output from the approaches is scaled back to real creep increments and accumulation is calculated over time. The effectiveness of the approaches is measured in terms of its ability to follow true pattern of creep accumulation.

Furthermore, training and validation loss histories were monitored during the training process in order to identify over- or underfitting. An approach is considered to be stable if the characteristics of both losses over the epochs almost coincide and converge.

### 3.4. Hyperparameter tuning

To optimize the hyperparameters of MLP, a simple manual grid search strategy is used. The best hyperparameters for the model trained with an amount of 100 % training data, corresponding to approximately 2225 h of temperature profile, are presented in Table 3. Because the MLP and its hyperparameters are quite well understood in the literature, we want to concentrate more on the LSTM model specific parameters in the following. Only one specific point should be mentioned, namely a callback method to stop training early if there is no decrease in validation loss within 100 epochs. For this reason, the number of epochs is adaptive for the MLP.

Both approaches, MLP and LSTM, minimize the losses using the Adam optimizer [21] with a defined initial learning rate.

For the LSTM architecture, rather small models of either 6 memory cells in the first and 4 memory cells in the second LSTM block, or 5 memory cells in each block, turned out to perform best. These models both feature slightly below 400 trainable parameters, which is significantly lower than the number of training samples for most sequence lengths and therefore counteracts overfitting. These models were trained for 500 epochs, saving the best result that occurred during the training process, which was usually located between 250 and 500 epochs. The rather small batch size of 25 for the LSTM model was chosen empirically and yielded good training results. The hyperparameters where identified using a semi-automatic parameter study, involving multiple small grid searches. In order to account for the randomness of weight initialization, the best model of three random initializations was determined for each hyperparameter combination.

The prediction accuracy of the LSTM network is furthermore highly dependent on the sequence length. Two major effects are influenced by it. A higher sequence length leads to extreme examples being less common in the training and test data, reducing the target range of creep strain increments, which facilitates the training process. At the same time, a higher sequence length reduces the number of samples in the training dataset, which generally leads to

**Table 3:** Hyperparameter best set with an amount of 100% training data

| Hyperparameter | value MLP | value LSTM |
|---|---|---|
| Optimization algorithm | Adam | Adam |
| Initial learning rate | 0.001 | 0.01 |
| Dropout rate | 0.15 | no dropout |
| Batch size | 512 | 25 |
| Epochs | Adaptive max. of 1000 | Best of 500 |
| Hidden layers | 2 | not used |
| Neurons per layer | 200 | not used |
| Memory cells | not used | First block 6 second block 4 |
| Sequence length | not used | Set to 165 for comparability to MLP |
| Sequence overlap | not used | 0.75 |

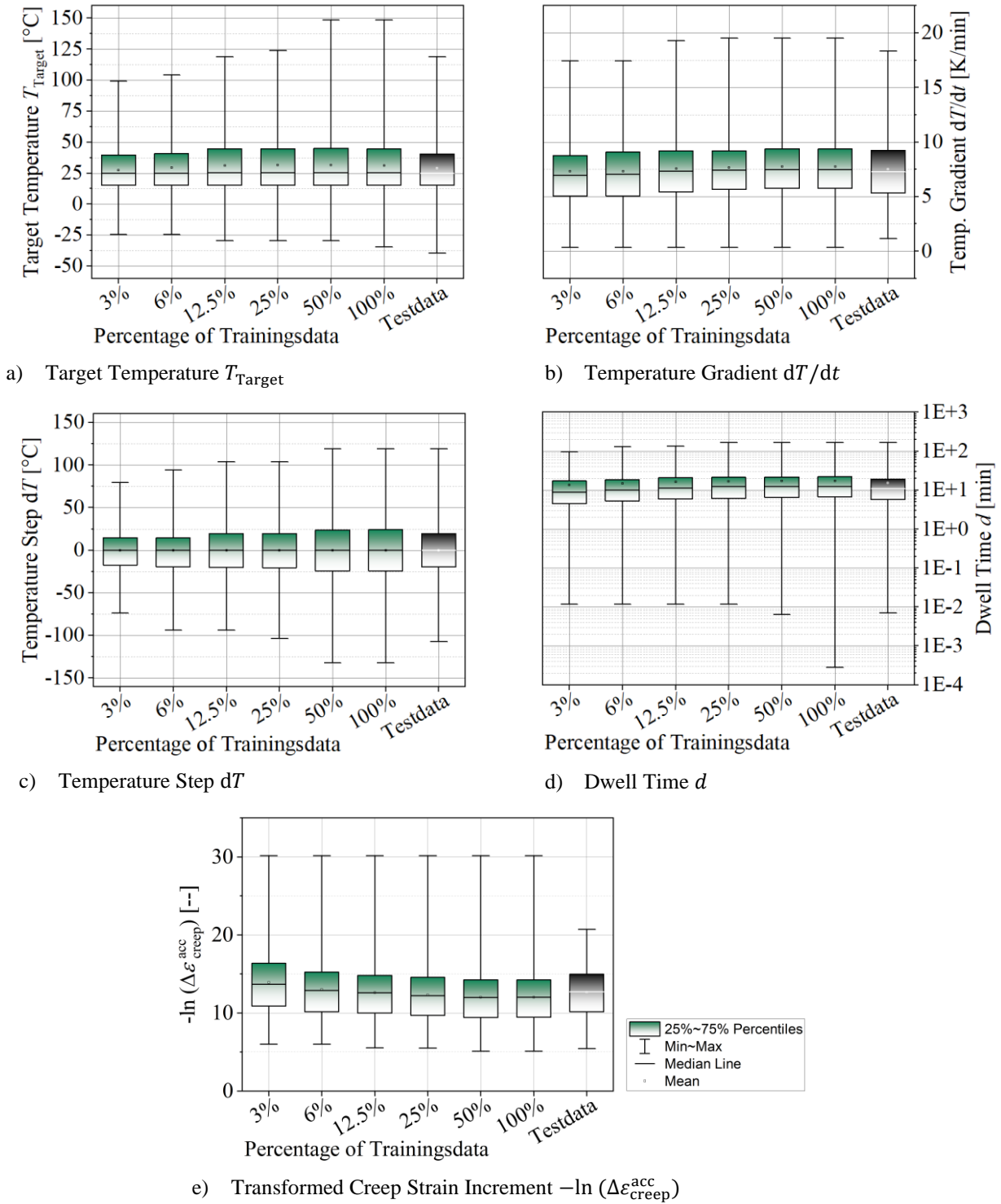

a) Target Temperature $T_{\text{Target}}$  
b) Temperature Gradient $dT/dt$  
c) Temperature Step $dT$  
d) Dwell Time $d$  
e) Transformed Creep Strain Increment $-\ln(\Delta\varepsilon_{\text{creep}}^{\text{acc}})$

**Figure 6:** Boxplots showing the distribution of the properties of the synthetic temperature profile for the different sized portions of the training dataset and the test dataset

worse predictions. In order to analyze how these effects interact, models where trained for a large range of sequence lengths and the results are displayed in Figure 7. Until a sequence length of about 100 min, the prediction quality drastically improves. Beyond a length of 500 min, the accuracy falls off again. To facilitate the comparison to the MLP model, a sequence length of 165 min was used for in Table 3 and further evaluations, as this is the longest sequence length of any individual half cycle in the training data.

### 3.5. Prediction results and comparison

After both approaches have been trained with the different length segments of the synthetic data set and for each of these trainings a separate optimization of the hyperparameters has been performed, the predictions of the models can now be compared with each other.

The prediction accuracy over all test samples is visualized by plotting the predicted values against the values from the FEA in Figure 8 for the LSTM approach and in Figure 9 for the MLP approach. Low dispersion from the center line in the plots indicates high degree of prediction accuracy, especially in the high creep region from 1E-4 to 0.01. Because in the present case the LSTM network is predicting based on a constant sequence length of 165 min, the predicted creep strain was generally larger, since the sequence length spans several half cycles. Conversely, the MLP calculates the creep strain increment for each individual half cycle. This also includes cycles with mostly insignificant creep effects, resulting in very low creep increments. The prediction over such a wide range from 1E-6 to 0.01 appears to be more difficult, so the MLP prediction deviation increases with lower increments. The cause is probably that despite the preventive efforts taken, such as the previous transfer of the creep expansions by means of the natural logarithm and the introduction of an adjusted loss function, smaller increments also have a smaller influence on the loss function and thus on the training of the weights. Here the LSTM network performed better with $R^2_{\text{LSTM}}$ score of 0.87 compared to MLP with $R^2_{\text{MLP}}$ score of 0.72.

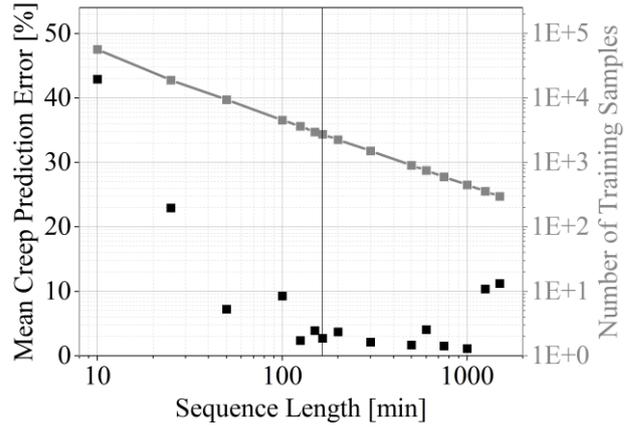

**Figure 7:** Dependence of prediction quality on subsequence length

Regardless of this, for the calculation of a RUL the quality of the predicted accumulated creep strain is important at the end. Figure 10 shows the predicted accumulated creep strains of the MLP and the LSTM compared to the FE solution. As it turns out, both approaches perform very well in predicting the test data set (20 % of the total profile) when trained on the full training data set (80 % of the total profile and approximately 2225 h of temperature profile). They are able to follow slow and constant increases as well as step changes in the creep strain.

The second question was, which amount of training data is necessary for the different approaches to deliver acceptable results. To answer this question, Figure 11 shows the averaged relative error according to Equation (12) plotted against the proportion of training data used. As can be seen, the prediction quality is very similar for both approaches when using the same amount of training data. In general, larger amounts of training data tend to lead to better predictions and smaller relative errors, but even with relatively small data sets of about 350 h of temperature profile (12.5 % of the available training data), both approaches deliver prediction errors of less than 10 %.

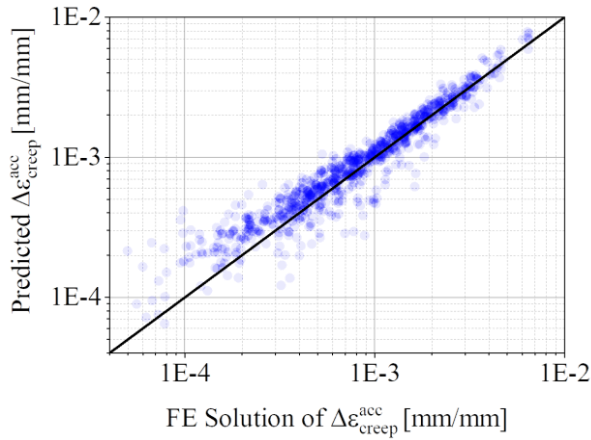

**Figure 8:** Prediction accuracy of LSTM network over all test samples, $R^2$ score: 0.874

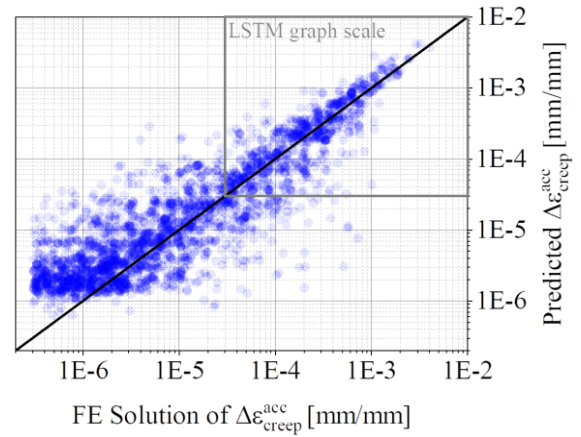

**Figure 9:** Prediction accuracy of MLP over all test samples, $R^2$ score: 0.720

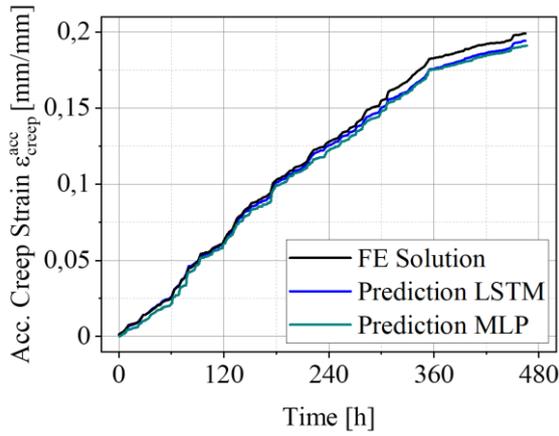

Figure 10: Visualization of creep strain prediction for the test data of MLP and LSTM model both trained with 100 % amount of training data

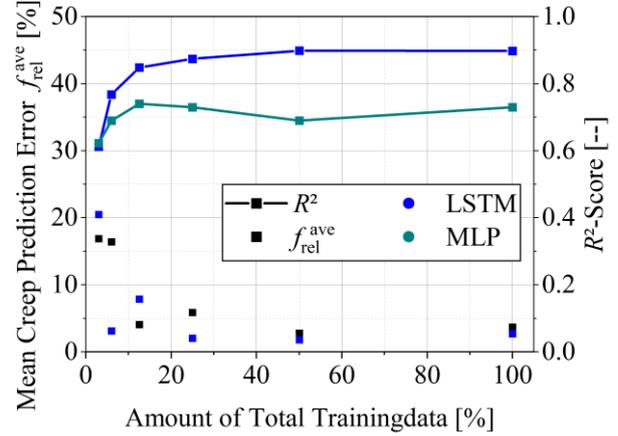

Figure 11: Dependence between the amount of training data used and the relative error calculated with equation (12)

## 4. Conclusions

The authors compared two real time capable, data driven approaches for predicting creep strain increments in solder joints of electronic components based on temperature profiles. The functionality of the approaches was demonstrated on synthetic data of the creep strain accumulation during thermal cycling serving as reference.

A synthetic, but for automotive applications realistic, temperature profile was generated, based on normal distributed predefined basic properties of thermal halfcycles. A finite element model of a bipolar chip resistor in SMT configuration was then used to enrich the synthetically data with creep strains. This reference data depending of the temperature profile and the corresponding creep strains, with a total length of approximately 2780 h was later separated into segments of different length. This allowed to quantitatively demonstrate the accuracy and potential of the investigated ML-approaches depending on the amount of provided training data. The statistical analyses of this segmented reference data sets ensure the normal distribution and with that the realistic behavior, showed that the definition of the temperature profile worked quite well. The very large range of the creep strain increments from 8E-14 to 6E-3 would have led to a misweighting of the data points during the training. To avoid this, a transformation by using the natural logarithm was done during data preparation.

The data driven approaches chosen, namely a MLP and a LSTM network, were briefly introduced in terms of structure, loss functions and hyperparameters. The hyperparameters were optimized and the parameters with the best performances were mentioned. A special focus was taken on the sequence length and its influence on the prediction error of the LSTM network because it affects the learning behavior in two opposite manners. On the one hand, a higher sequence length facilitates the training process because it leads to a kind of averaging of the creep strain increments, resulting in fewer extreme values. On the other hand, a higher sequence length reduces the number of samples in the training data set, which generally leads to weaker predictions. For the reported data, an optimum was found between 100 and 500 min long sequences and a value of 165 min was used for the further evaluations, as this is the longest sequence length of any individual half cycle in the training data.

It was quite surprising for the authors that the LSTM network, which is more complex in terms of its structure, shows a similarly low demand for training data as the MLP. The expectation that the MLP would be able to deliver better predictions for only few available training data, while the LSTM network would only deliver good predictions with significantly more training data and then also be more accurate than the MLP was not confirmed. Both approaches achieve prediction errors for the given synthetic data set smaller than 10 % by using 350 h training data length (12.5 % of the available training data) and smaller than 5 % by using 1400 h training data length (50 % of the available training data), respectively. Thus, both approaches have their relevance independent of the available amounts of training data. Therefore, the decision whether to use one of the two approaches shifts to the type of input variables available and the practicability of the value-to-value or sequence-to-value approach.

**Acknowledgments**

This research is funded by the European Regional Development Fund (ERDF) and co-financed by tax funds based on the budget approved by the members of the Saxon State Parliament. The authors are grateful to the Centre for Information Services and High Performance Computing of the TU Dresden for providing its facilities for high throughput calculations.

**References**

1. Reynell, M., "Advanced thermal analysis of packaged electronic systems using computational fluid dynamics techniques," *Computer-Aided Engineering Journal*, Vol. 7, Nr. 4 (1990), pp. 104–106.
2. Law, R.C. and Azid, I.A., "Application of artificial neural network in thermal and solder joint reliability

analysis for stacked dies LBGA," *Proc. 33rd IEEE/CPMT International Electronics Manufacturing Technology Conference (IEMT)*, 2008, pp. 1-7.
3. Lee, C. and Kwon, D., "A similarity based prognostics approach for real time health management of electronics using impedance analysis and SVM regression," *Microelectronics Reliability*, Vol. 83 (2018), pp. 77-83.
4. Lee, W., Nguyen, L. and Selvaduray, G., "Solder joint fatigue models: review and applicability to chip scale packages," *Microelectronics Reliability*, Vol. 40 (2000), pp. 231-244.
5. Syed, A., "Accumulated creep strain and energy density based thermal fatigue life prediction models for SnAgCu solder joints," *Proc. 54th Electronic Components and Technology Conference*, 2004, pp. 737-746.
6. Schwerz, R., Roellig, M. and Wolter, K.-J., "Reliability analysis of encapsulated components in 3D-circuit board integration," *Proc. 19th International Conference on Thermal, Mechanical and Multi-Physics Simulation and Experiments in Microelectronics and Microsystems (EuroSimE)*, 2018, pp. 1-12, doi: 10.1109/EuroSimE.2018.8369913.
7. Metasch, R., Roellig, M., Kabakchiev, A., Metais, B., Ratchev, R., Meier, K. and Wolter, K.-J., "Experimental investigation of the visco-plastic mechanical properties of a Sn-based solder alloy for material modelling in Finite Element calculations of automotive electronics," *Proc. 15th Thermal, Mechanical and Multi-Physics Simulation and Experiments in Microelectronics and Microsystems (EuroSimE)*, 2014, pp. 1–8.
8. Zahn, B., "Impact of ball via configurations on solder joint reliability in tapebased, chip-scale packages," *Proc. 52nd Electronic Components and Technology Conference (ECTC)*, 2002, pp. 1475–1483.
9. Feustel. F., FEM-Simulation der thermo-mechanischen Beanspruchung in Flip-Chip-Baugruppen zur Bewertung ihrer Zuverlässigkeit, Technische Universität Dresden (2002).
10. Lall, P., Islam, M.N., Singh, N., Suhling, J.C. and Darveaux, R., "Model for BGA and CSP reliability in automotive underhood applications," *IEEE Components and Packaging Technologies*, Vol. 27 (2004), pp. 585–593.
11 Chollet, F. (2015). Keras. GitHub repository. Retrieved from: https://github.com/fchollet/keras
12 TensorFlow Developers. (2021). TensorFlow (v2.6.0). Zenodo. https://doi.org/10.5281/zenodo.5181671
13 Goodfellow, I., Bengio, Y. and Courville, A., Deep learning, MIT press (2016).
14. LeCun, Y., Bengio, Y. and Hinton, G., "Deep learning," *nature*, Vol. 521, Nr. 7553 (2015), pp. 436-444.
15. Rumelhart, D.E., Hinton, G.E. and Williams, R.J., Learning internal representations by error propagation, California Univ. San Diego La Jolla Inst for Cognitive Science (1985).
16. He, K., Zhang, X., Ren, S., and Sun, J., "Delving deep into rectifiers: Surpassing human-level performance on imagenet classification," *Proc. International Conference on Computer Vision*, 2015, pp. 1026-1034.
17. Hochreiter, S., and Schmidhuber, J., "Long Short-Term Memory," *Neural Computation*, Vol. 9, Nr. 8 (1997), pp. 1735-1780.
18. Gers, F.A., Schmidhuber, J. and Cummins, F., "Learning to forget: continual prediction with (LSTM)," *Proc IET Conference*, 1999, pp. 850-855.
19. Bergstra, J. and Bengio, Y., "Random search for hyper-parameter optimization," *Journal of machine learning research*, Vol. 13, Nr. .2 (2012).
20. Bishop, C.M., Neural networks for pattern recognition, Oxford university press (1995).
21. Kingma, D.P. and Ba, J., "Adam: A method for stochastic optimization," *arXiv*, (2014), preprint arXiv:1412.6980

**Appendix A**

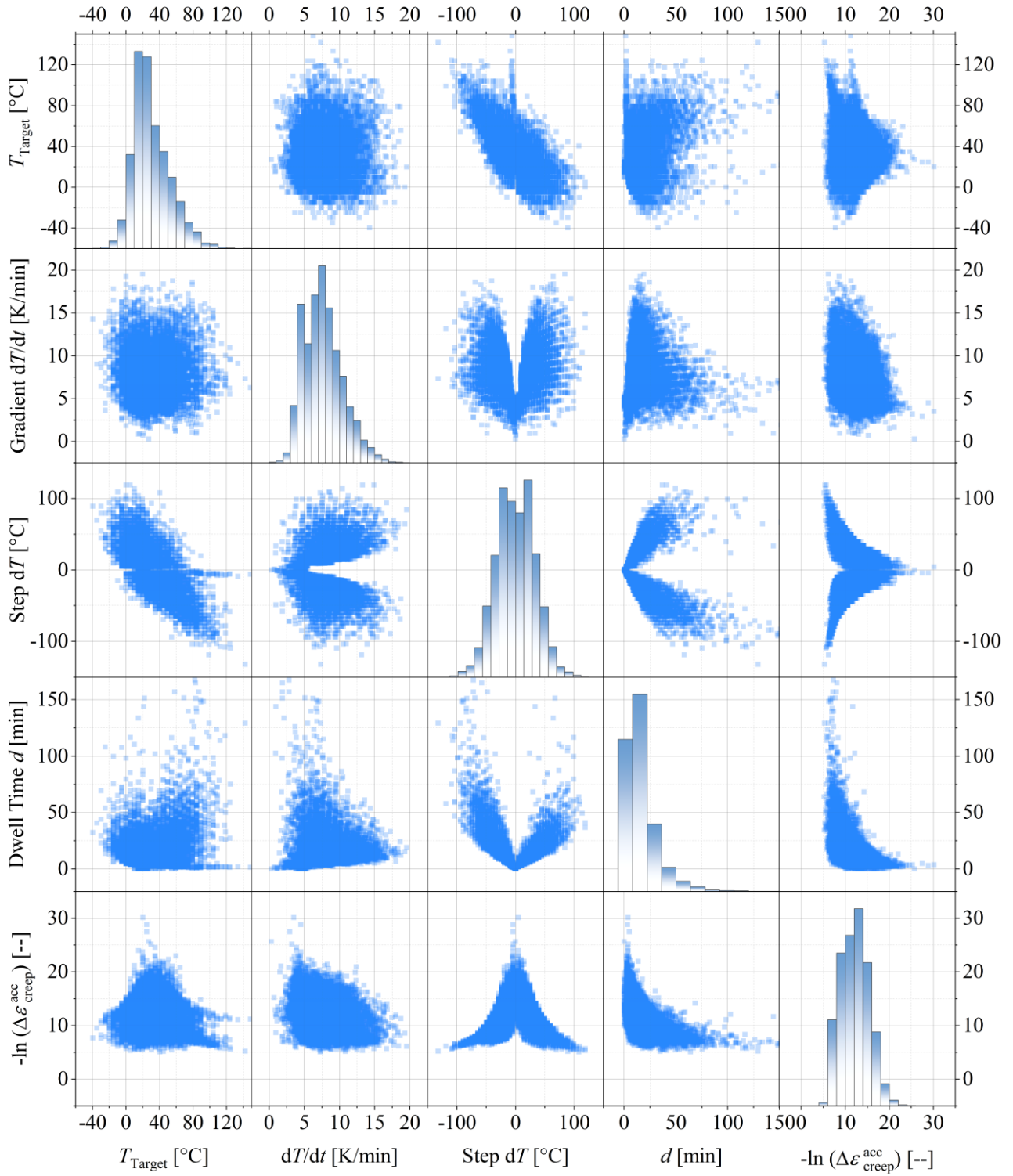

**Figure A-1:** Pair plot showing a normal distribution and the bivariate correlations of the synthetic data set